\documentclass{Interspeech}

\interspeechcameraready 

\usepackage[table,xcdraw,x11names]{xcolor}
\usepackage{caption}
\usepackage{subcaption}
\usepackage{multirow}
\newcommand{\dataset}{\textsc{Rasmalai}}
\newcommand{\model}{\textsc{IndicParlerTTS}}

\definecolor{darkgreen}{RGB}{0,128,0}
\title{\dataset~: \underline{R}esources for \underline{A}daptive \underline{S}peech \underline{M}odeling in Indi\underline{A}n \underline{L}anguages with \underline{A}ccents and \underline{I}ntonations}

\author[affiliation={1}]{Ashwin}{Sankar}
\author[affiliation={2}]{Yoach}{Lacombe}
\author[affiliation={1}]{Sherry}{Thomas}
\author[affiliation={1}]{Praveen}{Srinivasa Varadhan}
\author[affiliation={2}]{Sanchit}{Gandhi}
\author[affiliation={1}]{Mitesh}{M Khapra}

\affiliation{AI4Bharat, WSAI}{Indian Institute of Technology Madras}{India} 
\affiliation{}{Huggingface}{France}
\email{}
\keywords{speech synthesis, text-prompted, expressive}

\usepackage{comment}

\usepackage{booktabs}

\begin{document}

\maketitle

\begin{abstract}
We introduce \dataset, a large-scale speech dataset with rich text descriptions, designed to advance controllable and expressive text-to-speech (TTS) synthesis for 23 Indian languages and English. It comprises 13,000 hours of speech and 24 million text-description annotations with fine-grained attributes like speaker identity, accent, emotion, style, and background conditions. Using \dataset, we develop \model, the first open-source, text-description-guided TTS for Indian languages. Systematic evaluation demonstrates its ability to generate high-quality speech for named speakers, reliably follow text descriptions and accurately synthesize specified attributes. Additionally, it effectively transfers expressive characteristics both within and across languages. \model~consistently achieves strong performance across these evaluations, setting a new standard for controllable multilingual expressive speech synthesis in Indian languages.

\end{abstract}

\section{Introduction}

Text-to-Speech (TTS) synthesis has evolved beyond generating speech from a fixed set of voices using high-quality recordings, with recent advancements exploring speech- and text-prompted TTS systems. Speech-prompted TTS~\cite{peng2024voicecraft, li2023styletts, ju2024naturalspeech, microsoft2024E2} synthesizes speech by replicating the style and speaker characteristics from a provided audio prompt. In contrast, text-prompted TTS~\cite{leng2024prompttts, vyas2023audiobox, kawamura2024librittsp, lyth2024natural, jiang2024megatts} generates speech based on textual descriptions that can specify various attributes such as pitch, speed, noise levels, gender, and emotional expression. It is more user-friendly, as it removes the dependency on the availability of speech samples, allowing users to specify their desired speech characteristics via text. Training such models requires datasets containing triplets of \{\textit{audio}, \textit{transcript}, \textit{description}\}, where the \textit{description} guides the synthesis model to produce controlled and expressive speech with the specified attributes.

Much of the progress in text-prompted TTS has been limited to English and a few high-resource languages, while Indian languages  have lagged behind due to the lack of suitable datasets. While several efforts~\cite{baby2016resources, srinivasavaradhan2024rasa, iiith2024limmits} have introduced studio-recorded TTS datasets, their scale remains limited. Some TTS datasets~\cite{srinivasavaradhan2024rasa} provide detailed annotations for emotions and speaking styles, while others~\cite{sankar2024indicvoicesr}, though not recorded in controlled environments, offer greater scale, speaker diversity, and broader linguistic and acoustic coverage. These datasets vary in size, quality, and style diversity but none provide structured text descriptions that explicitly characterize speaker voice, style, environment and expressive attributes. This gap highlights the need for a dataset that integrates linguistic diversity, large-scale speech data, and rich textual descriptions to enable text-prompted TTS for Indian languages.

To address this gap, we propose generating text-based speech descriptions using Large Language Models (LLMs) by leveraging attribute information in the form of structured tags. Specifically, we aggregate multiple existing TTS datasets containing \{\textit{audio}, \textit{transcript}\} pairs and extract three categories of attributes: (i) metadata attributes such as age, gender, and speaker identity, where available; (ii) stylistic attributes, including Ekman emotions when annotated; and (iii) acoustic features such as pitch~\cite{morrison2023cross}, speaking rate, and clarity index~\cite{lavechin2023brouhaha, Bredin2020}, derived using speech processing libraries. These attributes are then provided as input to an LLM to generate detailed text descriptions, creating multiple text prompts per utterance. This methodology is applied to a diverse set of \{\textit{audio}, \textit{transcript}\} datasets spanning multiple Indic languages as well as English with varied accents. The resulting dataset, \dataset, comprises over 13,000 hours of transcribed speech, with each utterance annotated with three distinct text-based descriptions of varying specificity and robustness (see Table \ref{tab:example}). Additionally, these descriptions are translated into native languages using automated translation techniques \cite{ai4bharat2023it2}, further enhancing the dataset's inclusivity for multilingual text-prompted TTS.

\begin{table}[]
\centering
\fontsize{8pt}{10pt}\selectfont
\caption{Example showcasing (a) Descriptive, (b) Concise, and (c) Attribute Robust captions, for a set of speech attributes.}
\begin{tabular}{>{\centering\arraybackslash}p{0.25\linewidth}| p{0.65\linewidth}}
\toprule
\textbf{Attributes} & \textbf{Captions} \\
\midrule
\multirow{3}{=}[-1em]{\centering Jaya, female speaker, \\ Slightly close sounding, \\ High pitch, Expressive tone, \\ Slightly fast pace, \\ Great speech quality, \\ Speaking style is Anger} 
    & \textbf{(a)} Jaya, a female speaker, delivers speech in a slightly roomy environment with a high-pitched, expressive tone. She speaks at a slightly fast pace, with excellent overall speech quality. The intended style is anger. \\
\cmidrule(lr){2-2}
    & \textbf{(b)} Jaya, a female speaker, delivers high-pitched, expressive speech in a slightly enclosed environment at a fast pace, with excellent quality and an angry tone. \\
\cmidrule(lr){2-2}
    & \textbf{(c)} Jaya's angry tone, with a sharp voice, echoes with exceptional quality in a moderately reverberant environment. \\
\bottomrule
\end{tabular}
\label{tab:example}
\end{table}

Using \dataset, we train \model, the first multilingual text-prompted TTS system covering 24 Indian languages. We thoroughly evaluate its performance through multiple systematic assessments. First, we assess its ability to function as a general high-quality TTS system by synthesizing natural speech in the voice of named high-quality speakers, as specified in the text prompts. Using MUSHRA tests, we find that the synthesized speech exhibits high naturalness and strong fidelity to the intended speaker. Second, we evaluate the model's instruction-following capability by determining whether the generated speech accurately reflects the descriptions provided in the text prompts. Our results show that \model~effectively adheres to the specified characteristics. Third, we examine the model’s ability to generate expressive speech for speakers whose training data includes expressive variations. Our analysis shows that the model successfully synthesizes the intended emotions with very high accuracy. Fourth, we analyze the model’s capacity to synthesize expressive speech for speakers within a language even when no expressive data for that speaker was seen during training. We find that the model generalizes well in such cases, successfully performing style transfer to match the styles specified in the descriptions. Finally, we test its cross-lingual generalization capability by evaluating if the model can synthesize expressive speech in language $L_2$ for a speaker in language $L_1$, even in cases where no expressive speech data was available for that speaker in either language. \model~consistently achieves strong performance across all these evaluations, demonstrating its robustness and adaptability. To support further research in multilingual text-prompted TTS, we will release all models and code to the community.

\section{\dataset~: An Annotated Corpus for Controllable Multilingual TTS}
Below we describe (i) existing TTS datasets from which \{audio, text\} pairs were collated (ii) our approach for generating text descriptions for these audios and (iii) statistics of our dataset.

\subsection{Collating existing TTS datasets}
We first collate \{\textit{audio}, \textit{text}\} pairs from multiple existing Indian and English language datasets, as mentioned below:

\noindent\textbf{RASA \cite{srinivasavaradhan2024rasa}.}  ~A diverse studio-quality speech dataset that includes neutral readings, 
expressive speech with six basic Ekman emotions, and command-based interactions from platforms such as Alexa, UMANG, and DigiPay. It also includes spontaneous conversations, news readings, and audiobook narrations. Each utterance is annotated with style, emotion labels, and speaker identity. It comprises 20 speakers across 13 languages, totaling around 400 hours of speech.

\noindent\textbf{IndicVoices \cite{javed2024indicvoices}.}  A large-scale Indian speech dataset with over 7,200 hours of read, extempore, and conversational speech from 16,237 speakers across 22 languages. Originally designed for training ASR systems, it captures diverse acoustic, linguistic, and stylistic variations across various domains with recordings done in natural environments.

\noindent\textbf{IndicVoices-R \cite{sankar2024indicvoicesr}.} A 4,000-hour subset of IndicVoices, restored using  speech enhancement, with high-quality speech from 10,000+ speakers across 22 languages.



\noindent\textbf{IndicTTS \cite{baby2016resources} \& LIMMITS \cite{iiith2024limmits}.}  High-quality, studio-recorded neutral read speech datasets, with two speakers per language. LIMMITS covers 9 languages with 40 hours per speaker, while IndicTTS covers 13 languages, with 20 hours of native speech and 20 hours of English recordings per language.



\begin{table}[t]
    \fontsize{8pt}{9pt}\selectfont
    \centering
    \setlength{\tabcolsep}{2pt}
    \renewcommand{\arraystretch}{.9}
    \caption{Attributes coverage of different prompt-TTS Datasets}
    \label{tab:dataset_attributes}
    \begin{tabular}{lcccc}
        \toprule
        \textbf{Attributes} & \textsc{AudioBox} & \textsc{PromptTTS2} & \textsc{LibriTTS-P} & \dataset \\
        \midrule
        \textit{SNR} & \textcolor{darkgreen}{\checkmark} & \textcolor{red}{\texttimes} & \textcolor{darkgreen}{\checkmark} & \textcolor{darkgreen}{\checkmark} \\
        \textit{c50} & \textcolor{red}{\texttimes} & \textcolor{red}{\texttimes} & \textcolor{darkgreen}{\checkmark} & \textcolor{darkgreen}{\checkmark} \\
        \textit{Speaking Rate} & \textcolor{darkgreen}{\checkmark} & \textcolor{darkgreen}{\checkmark} & \textcolor{darkgreen}{\checkmark} & \textcolor{darkgreen}{\checkmark} \\
        \textit{PESQ} & \textcolor{darkgreen}{\checkmark} & \textcolor{red}{\texttimes} & \textcolor{darkgreen}{\checkmark} & \textcolor{darkgreen}{\checkmark} \\
        \textit{F0\textsubscript{mean}} & \textcolor{darkgreen}{\checkmark} & \textcolor{darkgreen}{\checkmark} & \textcolor{darkgreen}{\checkmark} & \textcolor{darkgreen}{\checkmark} \\
        \textit{F0\textsubscript{std}} & \textcolor{red}{\texttimes} & \textcolor{red}{\texttimes} & \textcolor{darkgreen}{\checkmark} & \textcolor{darkgreen}{\checkmark} \\
        \textit{Age} & \textcolor{darkgreen}{\checkmark} & \textcolor{red}{\texttimes} & \textcolor{darkgreen}{\checkmark} & \textcolor{darkgreen}{\checkmark} \\
        \textit{Speaker ID} & \textcolor{red}{\texttimes} & \textcolor{red}{\texttimes} & \textcolor{darkgreen}{\checkmark} & \textcolor{darkgreen}{\checkmark} \\
        \textit{Gender} & \textcolor{darkgreen}{\checkmark} & \textcolor{darkgreen}{\checkmark} & \textcolor{darkgreen}{\checkmark} & \textcolor{darkgreen}{\checkmark} \\
        \textit{Style} & \textcolor{darkgreen}{\checkmark} & \textcolor{darkgreen}{\checkmark} & \textcolor{darkgreen}{\checkmark} & \textcolor{darkgreen}{\checkmark} \\
        \textit{Accent} & \textcolor{darkgreen}{\checkmark} & \textcolor{red}{\texttimes} & \textcolor{red}{\texttimes} & \textcolor{darkgreen}{\checkmark} \\
        \textit{Env-tags} & \textcolor{darkgreen}{\checkmark} & \textcolor{red}{\texttimes} & \textcolor{red}{\texttimes} & \textcolor{darkgreen}{\checkmark} \\
        \textit{Multi-lingual} & \textcolor{red}{\texttimes} & \textcolor{red}{\texttimes} & \textcolor{red}{\texttimes} & \textcolor{darkgreen}{\checkmark} \\
        \bottomrule
    \end{tabular}
\end{table}

\noindent\textbf{GLOBE.} \cite{wang2024globe} A high-quality English speech corpus with 535 hours of speech from 23,519 speakers across 164 global accents. It refines Common Voice data through rigorous filtering, to enhance speech quality. The corpus is sampled at \SI{24}{\kilo\hertz} and includes detailed metadata with accent labels.

\subsection{Generating textual descriptions}
For all the \{\textit{audio}, \textit{text}\} pairs in the above datasets, we generate descriptions using a two-stage approach as described below:

\noindent\textbf{Extracting attributes from the audio.} 
Building on the pipeline from \cite{lyth2024natural}, we extract key acoustic features, such as pitch, C50, SNR, and speaking rate, and incorporate PESQ to assess overall speech quality. When available, we utilize the general metadata attributes like age and gender, as well as stylistic attributes such as expression and speech style. To ensure structured and consistent text descriptions, the continuous features, like SNR and C50, are discretized into bins, and all attributes are categorized.

\noindent\textbf{Generating description from prompts using an LLM.} For each utterance, we provide \textsc{Llama-3.1-8B-Instruct} with the categorized attributes, and prompt it to generate three types of textual descriptions (see Table \ref{tab:example}) - (i) \textit{Descriptive Prompt}: a detailed summary covering all labeled attributes, (ii) \textit{Concise Prompt}: a brief description of the sample, and (iii) \textit{Attribute-Robust Prompt}: which omits selected attributes to improve model robustness. We also translate the Descriptive Prompts into the respective language using IndicTrans2~\cite{ai4bharat2023it2}, generating a \textit{Native Prompt} to improve accessibility for multilingual users. 

\begin{figure*}
    \centering
    \caption{Comparison of durations for \textsc{Rasmalai-pretrain}, \textsc{Rasmalai-finetune} and \textsc{IndicVoices-R} across 24 languages.}
    \includegraphics[width=\linewidth]{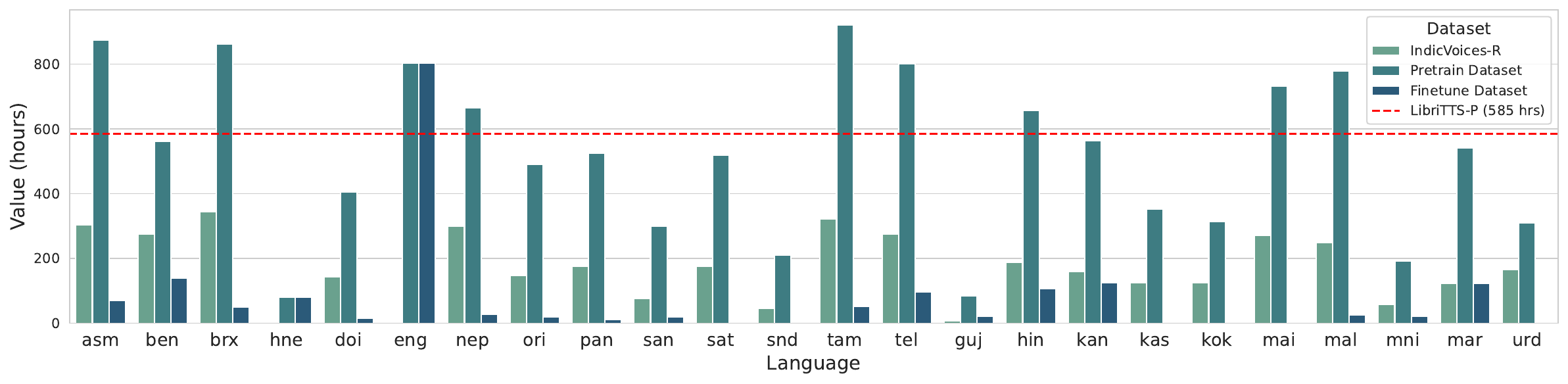}
    \label{fig:dataset_comparison}
\end{figure*}

\subsection{Dataset Statistics}
We provide two versions of the dataset: \textsc{Rasmalai-pretrain}, which includes the entire dataset, and \textsc{Rasmalai-finetune}, a studio-quality subset with named speakers. The finetuning set is designed for downstream tasks, focusing on improving speaker consistency and expression rendering. Figure \ref{fig:dataset_comparison} presents a breakdown of dataset duration per language, comparing \dataset~against IndicVoices-R. Notably, while prior open-source text-prompted TTS datasets~\cite{kawamura2024librittsp} for English contain only 585 hours of speech, our dataset substantially expands coverage to 23 Indian languages. Moreover, in the pre-training set, it surpasses the scale of English data for 8 languages, further enhancing multilingual speech synthesis resources. Additionally, as shown in Table~\ref{tab:dataset_attributes}, our dataset offers richer attribute coverage in comparison to existing text-prompted datasets, providing fine-grained control for accents, environments, and emotions. Overall, \dataset~ contains \textbf{13,000 hours} of speech data and \textbf{24 million text-description annotations}. 

\section{Experimental Setup}


\noindent \textbf{Model:} Building on Parler-TTS mini v1, we develop \model{} to support Indian languages by replacing the default tokenizer in Parler-TTS with an expanded Llama2 tokenizer, following the approach in ~\cite{mundra2024empirical}, enabling better subword segmentation in Indian languages.

\noindent \textbf{Training Setup:} We train \model~ using 32 H100 GPUs for 1.5 days, leveraging the open-sourced configuration from the original Parler-TTS implementation. We followed the hyperparameters outlined in the configuration, and replace the original scheduler with CosineLR with 3,000 warm-up steps, with an effective batch size of 256. 

\noindent\textbf{Data:} We pretrain the model on \dataset's pretraining set, which spans 24 languages and 13,000 hours of data, and further finetune it on the finetuning set, comprising high-fidelity data from 18 languages with 1,804 hours of speech. This enhances the speech quality while improving speaker consistency.

\noindent \textbf{Evaluation:} We evaluate our model against the previous state-of-the-art zero-shot TTS system, \textsc{IndicVoiceCraft}~\cite{sankar2024indicvoicesr} (\textsc{IndicVC}), across 13 Indian languages in the Rasa dataset, using MUSHRA and present the results in Tables \ref{tab:approaching_human_level} and \ref{tab:beats_sota}. Additionally, we report the consolidated CER, WER~\cite{javed2024indicvoices, norzoozi2024conformer}, MOS~\cite{noresqamos}, speaker similarity\footnote{\url{https://huggingface.co/microsoft/wavlm-base-plus-sv}}, and \textit{instruction adherence} in Table \ref{tab:auto_eval}. We evaluate instruction adherence by by re-annotating acoustic features of the synthesised samples using our pipeline up to the binning stage. We then compute IF-BLEU \cite{kawamura2024librittsp} by constructing prompts from binned values as comma-separated sequences and compare them to the original instruction from which this sample was synthesised.
We also report attribute-level accuracy in Table \ref{tab:instruction_following_accuracy}, which is calculated directly by comparing the binned values against the original attributes used to create the description. To evaluate emotion rendering, we conduct a perceptual emotion classification test with human raters and plot a confusion matrix. Finally, we extend our evaluation to a cross-lingual and cross-speaker setup to assess \model~for naturalness and expression generalization in a \textit{MUSHRA-like} evaluation. For all our evaluations, we employ 307 listeners who rated 5078 utterances, across multiple tests.
\section{Results}
We evaluate \model~on naturalness, style adherence, expressivity, and multilinguality.

\subsection{Approaching Human-level Synthesis on Seen Speakers}

Table \ref{tab:approaching_human_level} highlights languages where \model~approaches human-level naturalness, while Table \ref{tab:beats_sota} reports cases where our model significantly outperforms the previous best-performing system. As shown in Table \ref{tab:approaching_human_level}, \model~demonstrates high naturalness for seen expressive speakers in the Rasa test set covering 13 languages, achieving an average MUSHRA score of 81.7 compared to 89.7 for human speech. Furthermore, it approaches human-level speech synthesis in four languages: \textit{Bodo (brx), Maithili (mai), Marathi (mar),} and \textit{Telugu (tel)}, demonstrating its ability to generalize across language families while maintaining naturalness and high fidelity.  Table~\ref{tab:beats_sota} further illustrates our model's consistent improvement over the previous baseline on the Rasa test set, \textsc{IndicVC} with an average MUSHRA score of 81.7 versus 73.0. Notably, our model demonstrates significant improvements in \textit{Assamese (asm), Bengali (ben),} and \textit{Bodo (brx)}, elevating their MUSHRA scores from the Good (60–80) to the Excellent (80–100) range. Similarly, for \textit{Nepali (nep)}, the score improves from Fair (55.0) to Good (75.4), marking a substantial leap in perceived quality compared to \textsc{IndicVC} (Table \ref{tab:beats_sota}).

\begin{table}[]
\centering
\fontsize{8pt}{12pt}\selectfont
\setlength{\tabcolsep}{3pt} 
\renewcommand{\arraystretch}{1} 
\caption{Subjective evaluation of \model~on seen speakers in the \textit{Rasa-13} test set, with MUSHRA scores highlighting four languages reaching near-human naturalness.}
\label{tab:approaching_human_level}
\begin{tabular}{@{}lccccc@{}}
\toprule
\textit{\textbf{lang}} & \textit{\textbf{brx}} & \textit{\textbf{mai}} & \textit{\textbf{mar}} & \textit{\textbf{tel}} & \textit{\textbf{Rasa-13}} \\ \midrule
\textbf{Human}                  & 93.4 ± 1.8   & 87.7 ± 1.8   & 91.8 ± 1.1   & 92.4 ± 1.0   & 89.7 ± 1.8                \\ 
\textbf{Ours}                   & 85.4 ± 2.6   & 84.8 ± 1.9   & 88.0 ± 1.8   & 85.9 ± 2.3   & 81.7 ± 2.7                \\ \bottomrule
\end{tabular}

\end{table}

\begin{table}[]
\centering
\fontsize{8pt}{12pt}\selectfont
\setlength{\tabcolsep}{3pt} 
\renewcommand{\arraystretch}{1} 
\caption{MUSHRA scores highlighting languages where our model significantly surpasses previous best-performing system.}
\begin{tabular}{@{}lccccc@{}}
\toprule
\textit{\textbf{lang}} & \textit{\textbf{asm}} & \textit{\textbf{ben}} & \textit{\textbf{brx}} & \textit{\textbf{nep}} & \textit{\textbf{Rasa-13}} \\ \midrule
\textbf{IndicVC}       & 61.6 ± 4.9          & 73.2 ± 2.9          & 71.0 ± 3.2          & 55.0 ± 4.8          & 73.0 ± 3.4 \\
\textbf{Ours}          & 82.1 ± 2.4          & 83.4 ± 2.5          & 85.4 ± 2.6          & 75.4 ± 3.8          & 81.7 ± 2.7 \\ \bottomrule
\end{tabular}
\label{tab:beats_sota}
\end{table}

\subsection{Faithful Execution of Instructions}


In Table~\ref{tab:auto_eval}, we evaluate our model across key speech synthesis metrics, including intelligibility, naturalness, speaker consistency, and instruction adherence. The low CER (12\%) and WER (24\%) confirm that the model generates highly intelligible speech with minimal errors. A Noresqa-MOS score of 4.14, further indicates that the synthesized speech is natural and human-like. The speaker similarity score (0.95) demonstrates that our model is exceptionally consistent in preserving speaker identity across generated utterances. To assess instruction adherence, we report IF-BLEU, which reaches a high score of 93.18, showing the model's strong ability to follow given instructions accurately. 
Additionally, the attribute accuracy scores presented in Table~\ref{tab:instruction_following_accuracy} show strong performance in adhering to the specified attributes across most categories.
Together, these metrics highlight our model as a highly faithful executor of instructions. 
However, we note that PESQ accuracy is relatively lower, likely due to the finetuning dataset's focus on studio-quality recordings, limiting its ability to synthesize lower-quality speech.


\begin{table}[h]
\centering
\caption{Automatic evaluation results for \model.} 
\fontsize{8pt}{12pt}\selectfont
\begin{tabular}{@{}ccccc@{}}
\toprule
\textbf{CER (\%)} & \textbf{WER (\%)} & \textbf{MOS} & \textbf{S-SIM} & \textbf{IF-BLEU} \\ \midrule
12        & 24        & 4.14         & 0.95         & 93.18      \\ \bottomrule
\end{tabular}
\label{tab:auto_eval}
\end{table}

\begin{table}[h]
\centering
\fontsize{8pt}{12pt}\selectfont
\setlength{\tabcolsep}{3pt} 
\renewcommand{\arraystretch}{1} 
\caption{Table shows the per attribute accuracy (\%) of synthesized samples from the TTS subset of the test set.}
\begin{tabular}{@{}cccccc@{}}
\toprule
\textbf{c50} & \textbf{F0\textsubscript{std}} & \textbf{F0\textsubscript{mean}} & \textbf{SNR} & \textbf{Speaking Rate} & \textbf{PESQ}  \\ \midrule
99.26   & 96.61        & 86.2  & 96.83 & 97.12          & 80.44 \\ \bottomrule
\end{tabular}
\label{tab:instruction_following_accuracy}
\end{table}

\subsection{Evaluation of Expressive TTS}

\begin{figure}[!h]
    \centering
    \includegraphics[width=\linewidth]{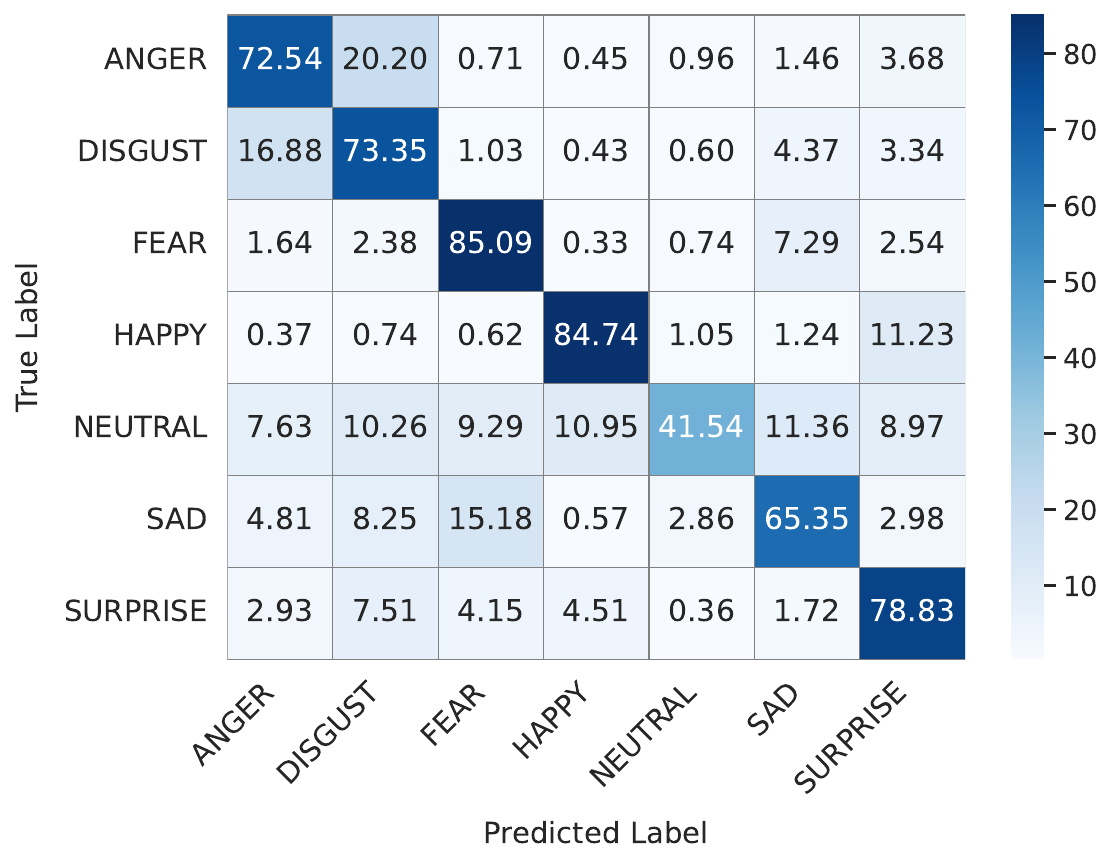}
    \caption{Confusion plot for Perceptual Emotion Classification}
    \label{fig:emotion_confusion}
\end{figure}

Our model exhibits strong emotion synthesis capabilities, as evaluated through a perceptual classification test. In this test, human listeners hear synthesized speech samples and classify them into one of seven target emotions. We report the accuracy of these human classifications for each emotion, providing insight into how well the synthesized speech conveys distinct emotional cues ( see Figure \ref{fig:emotion_confusion}). The confusion matrix shows that key emotions such as \textit{anger} (72.54\%), \textit{fear} (85.09\%), \textit{happy} (84.74\%), and \textit{surprise} (78.83\%) are recognized with high accuracy, indicating that the model effectively conveys distinct emotional cues in synthesized speech. However, some emotions exhibit higher confusion rates. Notably, \textit{sadness} (65.35\%) is misclassified as \textit{fear} 15.18\% of the time. This confusion likely arises from their shared acoustic features, such as lower pitch, slower speech rate, and reduced energy. 
Similarly, neutral speech is often confused with other emotions, as seen in Figure~\ref{fig:emotion_confusion}. 
This warrants further investigation to better understand the underlying factors contributing to this ambiguity.

\subsection{Zero-Shot Expressive Synthesis}
\noindent \model~demonstrates strong zero-shot expressive transfer, synthesizing expressive speech even for speakers with no expressive data in the training set. This is evident from the high MUSHRA scores (86.73) for Native speakers (expressing in their own language) in Table \ref{tab:x_speaker_expression}. Beyond native performance, our model also shows effective cross-lingual and cross-speaker expression transfer.  Proximal speakers (from related languages) achieve a MUSHRA score of 80.86, while Distal speakers (from unrelated languages) score 76.01. Notably, both groups generate expressive speech in a new language, despite no training data for these speakers in the target language or emotion. The slight degradation in expressiveness between Proximal and Distal speakers suggests that the model may struggle to generalize effectively to speakers from more distant language families, where differences in accent, dialect, and phonetic structure could contribute to lower expressivity scores.


\begin{table}[h]
    \centering
    \fontsize{8pt}{12pt}\selectfont
    \setlength{\tabcolsep}{3pt} 
    \renewcommand{\arraystretch}{1} 
    \caption{MUSHRA scores for zero-shot expressive synthesis across three different speaker groups.} 
    \label{tab:x_speaker_expression}
    \begin{tabular}{@{}ccc@{}}
        \toprule
        \textbf{Native}           & \textbf{Proximal}         & \textbf{Distal}        \\ \midrule
        86.73 $\pm$ 2.10 & 80.86 $\pm$ 3.28 & 76.01 $\pm$ 5.25 \\ \bottomrule
    \end{tabular}
\end{table}

\section{Related Work}

\noindent\textbf{Resources and Models for Text-Prompted TTS:} Recent works~\cite{leng2024prompttts, vyas2023audiobox, lyth2024natural, kawamura2024librittsp} have explored large-scale dataset annotation to improve TTS control through diverse speech attributes. However, most efforts are limited to English~\cite{leng2024prompttts, kawamura2024librittsp}, and 
cover only a narrow set of attributes with restricted domain diversity. Among these, only \cite{kawamura2024librittsp} publicly releases its dataset. \cite{lyth2024natural} introduces a language-agnostic approach for generating text annotations without relying on  human annotation or speech captioning models. 
In this work, we extend that pipeline to incorporate a broader range of attributes, and support more languages.

\noindent\textbf{Resources for Indian TTS:} Previous efforts~\cite{baby2016resources, srinivasavaradhan2024rasa, sankar2024indicvoicesr} have provided valuable datasets for Indian language TTS, supporting the development of single-speaker, expressive, and multilingual multi-speaker models. These resources have driven progress in zero-shot speaker and style-adaptive TTS. However, none of these datasets include the text annotations required for training text-prompted TTS models. 
\section{Conclusion}


We present \dataset, a dataset with rich textual descriptions for 13,000 hours of speech across 24 languages, covering diverse speakers, emotions, and styles, built using structured attribute extraction and LLM-based text generation. Using this dataset, we train \model, the first multilingual text-prompted TTS system for Indic languages. Our evaluations show that \model~produces highly natural, speaker-faithful speech, effectively follows text prompts, and synthesizes expressive speech with strong emotion accuracy. It also enables robust style transfer, generating expressive speech even for speakers without expressive training data, and demonstrates promising cross-lingual generalization. We release \dataset, along with our models and code, to drive progress in multilingual, expressive, and controllable TTS.

\bibliographystyle{IEEEtran}
\bibliography{mybib}

\end{document}